\documentclass[11pt,a4paper]{article}
\usepackage[hyperref]{acl2021}
\usepackage{times}
\usepackage{latexsym}

\usepackage{microtype}

\usepackage{amsfonts}
\usepackage{amsmath}
\usepackage{mathabx}
\usepackage{enumitem}
\usepackage[ruled,linesnumbered]{algorithm2e}
\usepackage{booktabs} 
\usepackage{threeparttable}
\usepackage{graphicx}
\usepackage{placeins}
\usepackage{subcaption}

\usepackage{url}
\usepackage{bbm}
\usepackage{amsthm}
\usepackage{multirow}

\newcommand{\mb}{\mathbf}
\newcommand{\mbs}{\boldsymbol}

\aclfinalcopy 

\title{Neural Bi-Lexicalized PCFG Induction}

\author{Songlin Yang$^{\clubsuit}$, Yanpeng Zhao$^{\diamondsuit}$, Kewei Tu$^{\clubsuit}$\thanks{\; Corresponding Author}\\
  $^{\clubsuit}$School of Information Science and Technology, ShanghaiTech University \\
  \textsuperscript{}{Shanghai Engineering Research Center of Intelligent Vision and Imaging}\\
  \textsuperscript{}{Shanghai Institute of Microsystem and Information Technology, Chinese Academy of Sciences}\\
    \textsuperscript{}{University of Chinese Academy of Sciences}\\
 $^{\diamondsuit}${ILCC, University of Edinburgh}\\
    {\tt \{yangsl,tukw\}@shanghaitech.edu.cn}\\
    {\tt yannzhao.ed@gmail.com}
 }

\date{}

\begin{document}
\maketitle
\begin{abstract}
Neural lexicalized PCFGs (L-PCFGs)~\citep{zhu-etal-2020-return} have been shown effective in grammar induction. However, to reduce computational complexity, they make a strong independence assumption on the generation of the child word and thus bilexical dependencies are ignored.
In this paper, we propose an approach to parameterize L-PCFGs without making implausible independence assumptions. Our approach directly models bilexical dependencies and meanwhile reduces both learning and representation complexities of L-PCFGs.  Experimental results on the English WSJ dataset confirm the effectiveness of our approach in improving both running speed and unsupervised parsing performance.
\end{abstract}

\section{Introduction}

Probabilistic context-free grammars (PCFGs) has been an important probabilistic approach to syntactic analysis~\citep{lari1990estimation,jelinek1992}.
They assign a probability to each of the parses admitted by CFGs and rank them by the plausibility in such a way that the ambiguity of CFGs can be ameliorated.
Still, due to the strong independence assumption of CFGs, 
vanilla PCFGs~\citep{Charniak1996} are far from adequate for highly ambiguous text.

A common premise for tackling the issue is to incorporate lexical information and weaken the independence assumption.
There have been many approaches proposed under the premise~\citep{magerman-1995-statistical,collins-1997-three,johnson-1998-pcfg,klein-manning-2003-accurate}.
Among them lexicalized PCFGs (L-PCFGs) are a relatively straightforward formalism~\citep{collins-2003-head}.
L-PCFGs extend PCFGs by associating a word, i.e., the lexical head, with each grammar symbol.
They can thus exploit lexical information to disambiguate parsing decisions and are much more expressive than vanilla PCFGs. However, they suffer from representation and inference  complexities.
For representation, the addition of lexical information 
greatly increases the number of parameters to be estimated and exacerbates the data sparsity problem during learning,
so the expectation-maximisation (EM) based estimation of L-PCFGs has to rely on sophisticated smoothing techniques and factorizations~\citep{collins-2003-head}. As for inference, the  CYK algorithm for L-PCFGs has a $O(l^5|G|)$ complexity, where $l$ is the sentence length and $|G|$ is the grammar constant. Although \citet{eisner-satta-1999-efficient} manage to reduce the complexity to $O(l^4|G|)$, inference with L-PCFGs is still relatively slow, making them less popular nowadays.





Recently, \citet{zhu-etal-2020-return} combine the ideas of factorizing the binary rule probabilities~\citep{collins-2003-head} and neural parameterization~\citep{kim-etal-2019-compound} and propose neural L-PCFGs (NL-PCFGs), achieving good results in both unsupervised dependency and constituency parsing.  Neural parameterization is the key to success, which facilitates informed smoothing~\cite{kim-etal-2019-compound}, reduces the number of learnable parameters for large grammars~\cite{chiu-rush-2020-scaling, yang-etal-2021-pcfgs} and facilitates advanced gradient-based optimization techniques instead of using the traditional EM algorithm~\cite{eisner-2016-inside}. However, \citet{zhu-etal-2020-return} oversimplify the binary rules to decrease the complexity of the inside/CYK algorithm in learning (i.e., estimating the marginal sentence log-likelihood) and inference. Specifically, they make a strong independence assumption on the generation of the child word such that 
it is only dependent on the nonterminal symbol. Bilexical dependencies, which have been shown useful in unsupervised dependency parsing ~\cite{han-etal-2017-dependency, yang-etal-2020-second}, are thus ignored.


To model bilexical dependencies and meanwhile reduce complexities, we draw inspiration from the canonical polyadic decomposition (CPD)~\cite{CPD} and propose a latent-variable based neural parameterization of L-PCFGs. \citet{cohen-etal-2013-approximate,  yang-etal-2021-pcfgs} have used CPD to decrease the complexities of PCFGs, and our work can be seen as an extension of their work to L-PCFGs. 
We further adopt the \emph{unfold-refold} transformation technique~\cite{ eisner-blatz-2007} to decrease complexities. By using this technique, we show that the time complexity of the inside algorithm implemented by \citet{zhu-etal-2020-return} can be improved from cubic to quadratic in the number of nonterminals $m$. The inside algorithm of our proposed method has a linear complexity in $m$ after combining CPD and unfold-refold.


We evaluate our model on the benchmarking Wall Street Journey (WSJ) dataset. Our model surpasses the strong baseline NL-PCFG~\cite{zhu-etal-2020-return} by 2.9$\%$ mean F1 and 1.3$\%$ mean UUAS under CYK decoding. When using the Minimal Bayes-Risk (MBR) decoding, our model performs even better. We provide an efficient implementation of our proposed model at \url{https://github.com/sustcsonglin/TN-PCFG}.

\section{Background}
\subsection{Lexicalized CFGs}
We first introduce the formalization of CFGs.
A CFG is defined as a 5-tuple $\mathcal{G} = (\mathcal{S},\mathcal{N}, \mathcal{P},\Sigma, \mathcal{R})$ where $\mathcal{S}$ is the start symbol, $\mathcal{N}$ is a finite set of nonterminal symbols, $\mathcal{P}$ is a finite set of preterminal symbols,\footnote{An alternative definition of CFGs does not distinguish nonterminals $\mathcal{N}$ (constituent labels) from preterminals $\mathcal{P}$ (part-of-speech tags) and treats both as nonterminals.} 
$\Sigma$ is a finite set of terminal symbols, and $\mathcal{R}$ is a set of rules in the following form:
\begin{align*}
	&S \rightarrow A   & \quad A \in \mathcal{N} \\
	&A \rightarrow B C, &\quad A \in \mathcal{N}, \quad B, C \in \mathcal{N} \cup \mathcal{P} \\ &T\rightarrow w, & \quad T \in \mathcal{P}, w \in \Sigma  
\end{align*}
$\mathcal{N}, \mathcal{P}$ and $\Sigma$ are mutually disjoint.
We will use `nonterminals' to indicate $\mathcal{N}\cup\mathcal{P}$ when it is clear from the context.

Lexicalized CFGs (L-CFGs)~\citep{collins-2003-head} extend CFGs by associating a word with each of the nonterminals:
\begin{align*}
	&S \rightarrow A[w_p]   &  A \in \mathcal{N} \\
	&A[w_p] \rightarrow B[w_p] C[w_q], & A \in \mathcal{N};  B, C \in \mathcal{N} \cup \mathcal{P} \\
	&A[w_p] \rightarrow C[w_q] B[w_p], & A \in \mathcal{N};  B, C \in \mathcal{N} \cup \mathcal{P}\\
	&T[w_p] \rightarrow w_p, & T \in \mathcal{P}
\end{align*}
where $w_p,w_q \in\Sigma$ are the headwords of the constituents spanned by the associated grammar symbols, and $p, q$ are the word positions in the sentence.
We refer to $A$, a parent nonterminal annotated by the headword $w_p$, as \emph{head-parent}.
In binary rules, we refer to a child nonterminal as \emph{head-child} if it inherits the headword of the head-parent (e.g., $B[w_p]$) and as \emph{non-head-child} otherwise  (e.g., $C[w_q]$).
A head-child appears as either the left child or the right child.
We denote the head direction by $D\in\{\curvearrowleft, \curvearrowright\}$, where $\curvearrowleft$ means head-child appears as the left child.

\subsection{Grammar induction with lexicalized probabilistic CFGs}

Lexicalized probabilistic CFGs (L-PCFGs) extend L-CFGs by assigning each production rule $r=A\rightarrow\gamma$ a scalar $\pi_r$ such that it forms a valid categorical probability distribution given the left hand side $A$.
Note that preterminal rules always have a probability of 1 because they define a deterministic generating process.

Grammar induction with L-PCFGs follows the same way of grammar induction with PCFGs.
As with PCFGs, we maximize the log-likelihood of each observed sentence $\mbs{w}=w_1,\ldots,w_l$:
\begin{align}
	\log p(\mbs{w}) = \log \sum_{t\in T_{\mathcal{G}_{L}}(\mbs{w})}  p(t)\,,
\end{align}
where $p(t) = \prod_{r\in t}\pi_r$ and $T_{\mathcal{G}_{L}}(\mbs{w})$ consists of all possible \emph{lexicalized} parse trees of the sentence $\mbs{w}$ under an L-PCFG $\mathcal{G}_{L}$.
We can compute the marginal $p(\mb{w})$ of the sentence by using the inside algorithm in polynomial time.
The core recursion of the inside algorithm is formalized in Equation~\ref{eq:inside_naive}.
It recursively computes the probability $s_{i, j}^{A, p}$ of a head-parent $A[w_p]$ spanning the substring $w_i,\ldots,w_{j-1}$ ($p\in[i, j-1]$).
Term A1 and A2 in Equation~\ref{eq:inside_naive} cover the cases of the head-child as the left child and the right child respectively.

\subsection{Challenges of L-PCFG induction}
\begin{figure*}[tb]
	\centering
	\includegraphics[width=0.75\linewidth]{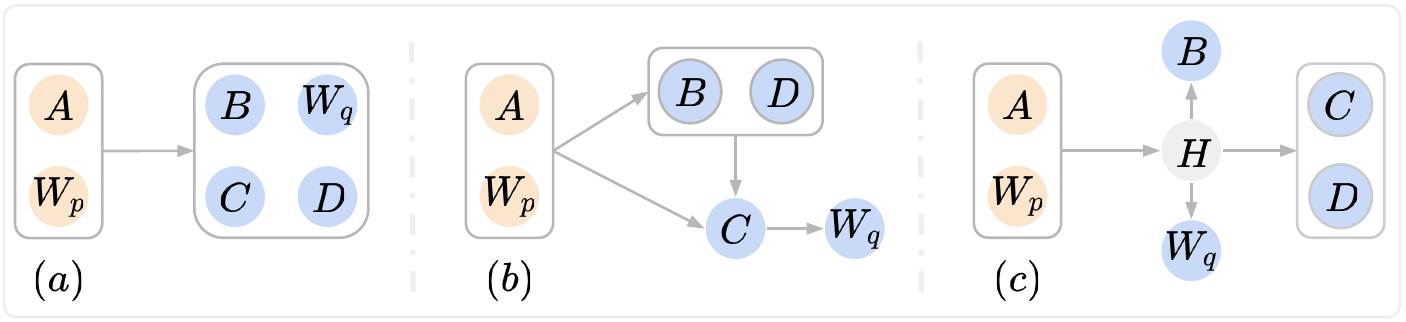}
	\caption{(a) The original parameterization of L-PCFGs. (b) The parameterization of \citet{zhu-etal-2020-return}: $W_q$ is independent with $B,D,A,W_p$ given C. (c) Our proposed parameterization. 
	We slightly abuse the Bayesian network notation by grouping variables. In the standard notation, there would be arcs from the parent variables to each grouped variable as well as arcs between the grouped variables.}
	\label{fig:bayesnet}
\end{figure*}

The major difference between L-PCFGs from vanilla PCFGs is that they use word-annotated nonterminals,
so the nonterminal number of L-PCFGs is up to $|\Sigma|$ times the number of nonterminals in PCFGs.
As the grammar size is largely determined by the number of binary rules 
and increases approximately in cubic of the nonterminal number,
representing L-PCFGs has a high space complexity $\mathcal{O}(m^{3}|\Sigma|^2)$ ($m$ is the nonterminal number).
Specifically, it requires an order-6 probability tensor for binary rules with each dimension representing $A$, $B$, $C$, $w_p$, $w_q$, and head direction $D$, respectively.
With so many rules,
L-PCFGs are very prone to the data sparsity problem in rule probability estimation. \citet{collins-2003-head} suggests factorizing the binary rule probabilities according to specific independence assumptions,
but his approach still relies on complicated smoothing techniques to be effective. 


The addition of lexical heads also scales up the computational complexity of the inside algorithm by a factor $\mathcal{O}(l^2)$ and brings it up to $\mathcal{O}(l^{5} m^{3})$.
\citet{eisner-satta-1999-efficient} point out that,
by changing the order of summations in Term A1 (A2) of Equation~\ref{eq:inside_naive},
one can cache and reuse Term B1 (B2) in Equation~\ref{eq:inside_eisner} and reduce the computational complexity to $\mathcal{O}(l^4m^2 + l^3m^3)$. This is an example application of unfold-refold as noted by \citet{ eisner-blatz-2007}. 
However, the complexity is still cubic in $m$, making it expensive to increase the total number of nonterminals.


\subsection{Neural L-PCFGs}

\citet{zhu-etal-2020-return} apply neural parameterization to tackle the data sparsity issue and to reduce the total learnable parameters of L-PCFGs. Considering the head-child as the left child (similarly for the other case), they further factorize the binary rule probability as:
\begin{align}
p(A&[w_{p}] \rightarrow B[w_{p}] C[w_{q}]) \nonumber\\
\label{eq:factor_zhu_left_prob}
&=p(B,\curvearrowleft, C | A, w_{p} )  p(w_{q} | C)\,. 
\end{align}

Bayesian networks representing the original probability and the factorization are illustrated in Figure \ref{fig:bayesnet} (a) and (b).
With the factorized binary rule probability in Equation~\ref{eq:factor_zhu_left_prob},
Term A1 in Equation~\ref{eq:inside_naive} can be rewritten as Equation~\ref{eq:inside_zhu_factor}.
\citet{zhu-etal-2020-return} implement the inside algorithm by caching Term C1-1 in Equation~\ref{eq:inside_zhu}, resulting in a time complexity $\mathcal{O}(l^4m^3 + l^3m)$, which is cubic in m.
We note that, 
we can use unfold-refold to further cache Term C1-2 in Equation~\ref{eq:inside_zhu} and reduce the time complexity of the inside algorithm to $\mathcal{O}(l^4m^2 + l^3m + l^2m^2)$,
which is quadratic in $m$. 


Although the factorization of Equation~\ref{eq:factor_zhu_left_prob} reduces the space and time complexity of the inside algorithm of  L-PCFG, it is based on the independence assumption that the generation of $w_q$ is independent of $A$, $B$, $D$ and $w_p$ given the non-head-child $C$.
This assumption can be violated in many scenarios and hence reduces the expressiveness of the grammar.
For example, suppose $C$ is Noun, then even if we know $B$ is Verb, we still need to know $D$ to determine if $w_q$ is an object or a subject of the verb, and then need to know the actual verb $w_p$ to pick a likely noun as $w_q$. 

\begin{table*}[!ht]
	\centering
	{\setlength{\tabcolsep}{.0em}
		\begin{tabular}{l}
			\toprule 
			\begin{minipage}{\linewidth}
				\vspace{-.5em}\small
				{\begin{flalign}
					\label{eq:inside_naive}
						\!\!\!\!\!\!\!s_{i,j} ^{A,p} 	=& \underbrace {\sum_{k=p+1}^{j - 1} \sum_{q=k}^{j - 1}\sum_{B, C}   s_{i, k}^{B,p} \cdot s_{k, j}^{C,q} \cdot p(A[w_{p}] \rightarrow B[w_{p}] C[w_{q}])}_{\text{Term A1}} +  \underbrace {\sum_{k=i+1}^{p}\sum_{q=i}^{k - 1}\sum_{B, C} s_{i, k}^{B,q} \cdot  s_{k, j}^{C,p} \cdot p(A[w_{p}] \rightarrow B[w_{q}] C[w_{p}])}_{\text{Term A2}} \\
					\label{eq:inside_eisner}
					=& \sum_{k=p+1}^{j - 1} \sum_{B} s_{i, k}^{B,p}  \underbrace{ \sum_{q=k}^{j - 1}\sum_{C}  s_{k, j}^{ C,q} \cdot p(A[w_{p}] \rightarrow B[w_{p}] C[w_{q}])}_{\text{Term B1}} +\sum_{k=i+1}^{p}  \sum_{C}   s_{k_2, j}^{C,p} \underbrace{ \sum_{q=i}^{k - 1}\sum_{B} s_{i, k}^{B,q} \cdot p(A[w_{p}] \rightarrow B[w_{q}] C[w_{p}])}_{\text{Term B2}}
					\end{flalign}}
				\vspace{-.5em}
			\end{minipage}\\
			\toprule 
			\begin{minipage}{\linewidth}
				\vspace{-.5em}\small
				{\begin{align}
					\label{eq:inside_zhu_factor}
					\text{Term A1}  &= \sum_{k=p+1}^{j - 1}\sum_{q=k}^{j - 1}\sum_{B, C}   s_{i, k}^{B,p} \cdot s_{k, j}^{C,q} \cdot \underbrace{p(B, \curvearrowleft, C | A,w_p) \cdot  p(w_{q}| C )}_{\text{factorization of } p(A[w_{p}] \rightarrow B[w_{p}] C[w_{q}])} \\
					\label{eq:inside_zhu}
					&= \sum_{k=p+1}^{j - 1}\sum_{B}  s_{i, k}^{B, p} \underbrace{\sum_{C} p(B, \curvearrowleft, C | A,w_p)   \underbrace{ \sum_{q = k}^{j - 1}  s_{k, j}^{C,q} \cdot  p(w_{q}| C )}_{\text{Term C1-1}}}_{\text{Term C1-2}} 
					\end{align}}
				\vspace{-.5em}
			\end{minipage}\\
			\toprule
			\begin{minipage}{\linewidth}
				\vspace{-.5em}\small
				{\begin{align}
					\label{eq:inside_ours_factor}
					\text{Term A1} &= \sum_{k=p+1}^{j - 1}\sum_{q=k}^{j - 1}\sum_{B, C}   s_{i, k}^{B,p} \cdot s_{k, j}^{C,q} \cdot\underbrace{\sum_{H} p(H | A, w_{p}) p(B | H)  p(C, \curvearrowleft | H)  p( w_{q} | H)}_{\text{factorization of } p(A[w_{p}] \rightarrow B[w_{p}] C[w_{q}])} \\
					\label{eq:inside_ours}
					&=  \sum_{H}  p(H |A, w_p) \sum_{k=p+1}^{j - 1} \underbrace{\vphantom{\sum_{q=k}^{j - 1}}\sum_{B} s_{i, k}^{B, p} p(B | H)}_{\text{Term D1-1}}  \underbrace{\sum_{q=k}^{j - 1}\sum_{C} s_{k, j}^{C,q} p (C \curvearrowleft | H) p( w_q | H)}_{\text{Term D1-2}}
					\end{align}}
				\vspace{-.5em}
			\end{minipage}\\
			\bottomrule
	\end{tabular}}
	\caption{
		\label{tab:inside_equations}
		Recursive formulas of the inside algorithm for \citet{eisner-satta-1999-efficient} (Equation~\ref{eq:inside_eisner}), \citet{zhu-etal-2020-return} (Equation~\ref{eq:inside_zhu_factor}-~\ref{eq:inside_zhu}), and our formalism (Equation~\ref{eq:inside_ours_factor}-~\ref{eq:inside_ours}), respectively. $s_{i, j}^{A, p}$ indicates the probability of a head nonterminal $A[w_p]$ spanning the substring $w_i,\ldots,w_{j-1}$, where $p$ is the position of the headword in the sentence.
	}
\end{table*}




\section{Factorization with latent variable}

Our main goal is to find a parameterization that removes the implausible independence assumptions of \citet{zhu-etal-2020-return} while decreases the complexities of the original L-PCFGs.


To reduce the representation complexity, we draw inspiration from the canonical polyadic decomposition (CPD). CPD factorizes an $n$-th order tensor into $n$ two-dimensional matrices. Each matrix consists of two dimensions: one dimension comes from the original $n$-th order tensor and the other dimension is shared by all the $n$ matrices. The shared dimension can be marginalized to recover the original $n$-th order tensor. From a probabilistic perspective, the shared dimension can be regarded as a latent-variable. In the spirit of CPD, we introduce a latent-variable $H$ to decompose the order-6 probability tensor $p(B, C, D, w_q | A, w_p)$. Instead of fully decomposing the tensor, we empirically find that binding some of the variables leads to better results. Our best factorization is as follows (also illustrated by a Bayesian network in Figure \ref{fig:bayesnet} (c)):

\begin{align}
&p(B, C, W_q, D | A, W_p) =  \\
&\sum_{H} p(H| A, W_p) p(B | H) p(C, D| H) p(W_q| H)\,.\nonumber
\end{align}
According to d-separation~\citep{dseparation}, 
when $A$ and $w_p$ are given, $B$, $C$, $w_q$, and $D$ are interdependent due to the existence of $H$.
In other words, our factorization does not make any independence assumption beyond the original binary rule.  The domain size of $H$ is analogous to the tensor rank in CPD and thus influences the expressiveness of our proposed model. 

Based on our factorization approach, the binary rule probability is factorized as
\begin{align}
\label{eq:factor_ours_left}
    \!\!\!&p(A[w_p]\rightarrow B[w_p] C[w_q]) = \\ \nonumber
&\sum_{H} p(H| A, w_p)p(B|H)p(C\curvearrowleft|H)p(w_q|H) \,,
\end{align}
and 
\begin{align}
\label{eq:factor_ours_right}
\!\!\!&p(A[w_p] \rightarrow B[w_q] C[w_p]) = \\\nonumber
&\sum_{H} p(H | A, w_p)p(C|H)p(B \curvearrowright|H)p(w_q|H)\,.
\end{align}
We also follow \citet{zhu-etal-2020-return} and factorize the start rule as follows.
\begin{align}\label{eq:factor_zhu_root}
p(S\rightarrow A[w_p]) 
&=p(A | S)p(w_p | A)\,.
\end{align}

\paragraph{Computational complexity:} Considering the head-child as the left child (similarly for the other case),
we apply Equation~\ref{eq:factor_ours_left} in Term A1 of Equation~\ref{eq:inside_naive} and obtain Equation~\ref{eq:inside_ours_factor}.
Rearranging the summations in Equation~\ref{eq:inside_ours_factor} gives Equation~\ref{eq:inside_ours},
where Term D1-1 and D1-2 can be cached and reused, which also uses the  unfold-refold technique. 
The final time complexity of the inside computation with our factorization approach is $\mathcal{O}(l^4d_H + l^2md_H)$ ($d_H$ is the domain size of the latent variable $H$), which is linear in $m$.

\paragraph{Choices of factorization:}
If we follow the intuition of CPD, then we shall assume that $B$, $C$, $D$, and $w_q$ are all independent conditioned on $H$. However, properly relaxing this strong assumption by binding some variables could benefit our model.
Though there are many different choices of binding the variables,
some bindings can be easily ruled out.
For instance, binding $B$ and $C$ inhibits us from caching Term D1-1 and Term D1-2 in Equation \ref{eq:inside_ours_factor} and thus we cannot implement the inside algorithm efficiently;
binding $C$ and $w_q$ leads to a high computational complexity because we will have to compute a high-dimensional ($m|\Sigma|$) categorical distribution. In Section \ref{section:abl_bind}, we make an ablation study on the impact of different choices of factorizations.

\paragraph{Neural parameterizations:}
We follow \citet{kim-etal-2019-compound} and \citet{zhu-etal-2020-return} and define the following neural parameterization:
\begin{align*}
p( A|S) &= \frac{\exp (\mathbf{u}_{S}^{\top} f_1(\mb{w}_{A}))}{  \sum_{A^{\prime} \in \mathcal{N}} \exp (\mathbf{u}_{S}^{\top} f_1(\mb{w}_{A^{\prime}})) } \,,\\
p(w|A) &= \frac{\exp (\mathbf{u}_{A}^{\top} f_{2}( \mb{w}_{w}))}{  \sum_{w^{\prime} \in \Sigma} \exp (\mathbf{u}_{A}^{\top} f_{2}(\mb{w}_{w'}))}\,, \\
p(B | H) &= \frac{\exp (\mathbf{u}_{H}^{\top} w_{B})}{  \sum_{B^{\prime} \in \mathcal{N}\cup \mathcal{P}} \exp (\mathbf{u}_{H}^{\top} \mb{w}_{B^{\prime}}) } \,,\\
p(w|H) &= \frac{\exp (\mathbf{u}_{H}^{\top} f_2 ( \mb{w}_{w}) )}{  \sum_{w^{\prime} \in \Sigma} \exp (\mathbf{u}_{H}^{\top} f_2 ( \mb{w}_{w^{\prime}})) } \,, \\
p( C \curvearrowleft | H) &=  \frac{ \exp (\mathbf{u}_{H}^{\top} \mb{w}_{C\curvearrowleft})}{ \sum_{C^{\prime} \in \mathcal{M}} \exp (\mathbf{u}_{H}^{\top} \mb{w}_{C^{\prime}})  }  \,,\\
p(C \curvearrowright |H) &=  \frac{ \exp (\mathbf{u}_{H}^{\top} \mb{w}_{C\curvearrowright})}{ \sum_{C^{\prime} \in \mathcal{M}} \exp (\mathbf{u}_{H}^{\top} \mb{w}_{C^{\prime}})  } \,, \\
p( H|A, w) &=  \frac{ \exp (\mathbf{u}_{H}^{\top} f_{4}([\mathbf{w}_{A}; \mathbf{w}_{w}]) ) }{\sum_{H^{\prime} \in \mathcal{H}} \exp (\mathbf{u}_{H^{\prime}}^{\top} f_{4}([\mathbf{w}_{A}; \mathbf{w}_{w}])) } \,,
\end{align*}
where $\mathcal{H} = \{H_{1},\ldots,H_{d_H}\}$, $\mathcal{M} = (\mathcal{N}\cup\mathcal{P}) \times \{\curvearrowleft, \curvearrowright\}$,
$\mb{u}$ and $\mb{w}$ are nonterminal embeddings and word embeddings respectively,
and $f_1(\cdot)$, $f_2(\cdot)$, $f_3(\cdot)$, $f_4(\cdot)$ are neural networks with residual layers~\citep{he2016residual} (Full parameterization is shown in Appendix.).


 


\section{Experimental setup}
\subsection{Dataset}
We conduct experiments  on the Wall Street Journal (WSJ) corpus of the Penn Treebank~\citep{marcus-etal-1994-penn}.
We use the same preprocessing pipeline as in \citet{kim-etal-2019-compound}.
Specifically, punctuation is removed from all data splits
and the top 10,000 frequent words in the training data are used as the vocabulary. 
For dependency grammar induction,
we follow~\cite{zhu-etal-2020-return} to use the Stanford typed dependency representation~\citep{de-marneffe-manning-2008-stanford}.

\subsection{Hyperparameters}
We optimize our model using the Adam optimizer with $\beta_1 = 0.75, \beta_2 = 0.999$, and learning rate $0.001$.
All parameters are initialized with Xavier uniform initialization. We set the dimension of all embeddings to 256 and the ratio of the nonterminal number to the preterminal number to 1:2. Our best model uses 15 nonterminals, 30 preterminals, and $d_H = 300$.
We use grid search to tune the nonterminal number (from 5 to 30) and domain size $d_H$ of the latent $H$ (from 50 to 500).

\subsection{Evaluation}
We run each model four times with different random seeds and for ten epochs. We train our models on training sentences of length $\le 40$ with batch size 8 and test them on the whole testing set. 
For each run, we perform early stopping and select the best model according to the perplexity of the development set. We use two different parsing methods: the variant of CYK algorithm ~\cite{eisner-satta-1999-efficient} and Minimum Bayes-Risk (MBR) decoding ~\cite{smith-eisner-2006-minimum}. \footnote{In MBR decoding, we use automatic differentiation ~\citep{eisner-2016-inside, rush-2020-torch} to estimate the marginals of spans and arcs, and then use the CYK and Eisner algorithms for constituency and dependency parsing, respectively.} For constituent grammar induction, we report the means and standard deviations of sentence-level F1 scores.\footnote{Following \citet{kim-etal-2019-compound}, we remove all trivial spans (single-word spans and sentence-level spans). Sentence-level means that we compute F1 for each sentence and then average over all sentences.} For dependency grammar induction, we report unlabeled directed attachment score (UDAS) and unlabeled undirected attachment score (UUAS). 

\section{Main result}             
We present our main results in Table \ref{tab:main_result}. Our model is referred to as \textbf{N}eural \textbf{B}i-\textbf{L}exicalized \textbf{PCFGs} (NBL-PCFGs).
We mainly compare our approach against recent PCFG-based models: neural PCFG (N-PCFG) and compound PCFG (C-PCFG) \cite{kim-etal-2019-compound}, tensor decomposition based neural PCFG (TN-PCFG) \cite{yang-etal-2021-pcfgs} and neural L-PCFG (NL-PCFG) \cite{zhu-etal-2020-return}.
We report both official result of \citet{zhu-etal-2020-return} and our reimplementation. 

We do not use the compound trick~\citep{kim-etal-2019-compound} in our implementations of lexicalized PCFGs because we empirically find that using it results in unstable training and does not necessarily bring performance improvements.



We draw three  key observations:
(1) Our model achieves the best F1 and UUAS scores under both CYK and MBR decoding. 
It is also comparable to the official NL-PCFG in the UDAS score.
(2) When we remove the compound parameterization from NL-PCFG, its F1 score drops slightly while its UDAS and UUAS scores drop dramatically. It implies that compound parameterization is the key to achieve excellent dependency grammar induction performance in NL-PCFG. 
(3) The MBR decoding outperforms CYK decoding. 


    Regarding UDAS, our model significantly outperforms NL-PCFGs in UDASs if compound parameterization is not used (37.1 vs. 23.8 with CYK decoding), showing that explicitly modeling bilexical relationship is helpful in dependency grammar induction. However, when compound parameterization is used, the UDAS of NL-PCFGs is greatly improved, slightly surpassing that of our model. We believe this is because compound parameterization greatly weakens the independence assumption of NL-PCFGs (i.e., the child word is dependent on C only) by leaking bilexical information via the global sentence embedding. On the other hand, NBL-PCFGs are already expressive enough and thus compound parameterization brings no further increase of their expressiveness but makes learning more difficult. 

\begin{table}[t!] 	 
	{\setlength{\tabcolsep}{.5em}
		\makebox[\linewidth]{\resizebox{\linewidth}{!}{%
				\begin{tabular}{llll}
					\toprule 
					\multirow{2}{*}{Model} & \multicolumn{3}{c} { WSJ } \\
					\cmidrule{2-4}
					& F1 & UDAS & UUAS   \\
										\midrule 
									\multicolumn{4}{c}{Official results}                                   \\
														\midrule 
					N-PCFG$^\star$& 50.8   \\
					C-PCFG$^\star$& 55.2  \\
					NL-PCFG$^\star$& 55.3 & \textbf{39.7} & 53.3 \\
					TN-PCFG$^\dagger$& 57.7 \\ 
					\midrule 

							\multicolumn{4}{c}{Our results}                                   \\
							\midrule
					NL-PCFG$^\star$ & 53.3$_{\pm 2.1}$ & 23.8$_{\pm 1.1}$ & 47.4$_{\pm 1.0}$   \\ 
										NL-PCFG$^\dagger$& 57.4$_{\pm 1.4}$ & 25.3$_{\pm 1.3}$ & 47.2$_{\pm 0.7}$   \\ 
                    
                    NBL-PCFG$^\star$  & 58.2$_{\pm 1.5}$ & 37.1$_{\pm 2.8}$ & 54.6$_{\pm 1.3}$  \\
                    
                    NBL-PCFG$^\dagger$
                    & \textbf{60.4}$_{\pm 1.6}$ & 39.1$_{\pm 2.8}$ & \textbf{56.1}$_{\pm 1.3}$ \\
					\midrule
				\multicolumn{4}{c}{For reference}                                   \\
			
					\midrule 
			        	
        				S-DIORA
        				& 57.6  \\					
                StructFormer & 54.0 & 46.2 & 61.6 \\ 				
					\midrule 
					Oracle Trees &84.3 &  \\
					\bottomrule
	\end{tabular}}}}
	\caption{\label{tab:main_result} Unlabeled sentence-level F1 scores, unlabeled directed attachment scores and unlabeled undirected attachment scores on the WSJ test data.  
		$^\dagger$ indicates using MBR decoding. 
		$^\star$ indicates using CYK decoding. Recall that the official result of \citet{zhu-etal-2020-return} uses compound parameterization while our reimplementation removes the compound parameterization. S-DIORA: \citet{drozdov-etal-2020-unsupervised}. StructFormer: \citet{shen2020structformer}.
		}
		\vskip -.1in
\end{table}

\section{Analysis}
  In the following experiments, we report results using MBR decoding by default. We also use $d_H=300$ by default unless otherwise specified.

\subsection{Influence of the domain size of $H$}
\label{sec:inf_h}
$d_H$ (the domain size of $H$) influences the expressiveness of our model. Figure \ref{fig:f1_perp_rank} illustrates perplexities and F1 scores with the increase of $d_H$ and a fixed nonterminal number of 10 (plots of UDAS and UUAS can be found in Appendix).  We can see that when $d_H$ is small, the model has a high perplexity and a low F1 score, indicating the limited expressiveness of NBL-PCFGs. When $d_H$ is larger than 300, the perplexity becomes plateaued and the F1 score starts to decrease possibly because of overfitting.

\subsection{Influence of nonterminal number}
\label{sec:inf_nt}
\begin{figure*}[ht!]
	\begin{subfigure}[t]{0.49\linewidth}
		\includegraphics[scale=0.48]{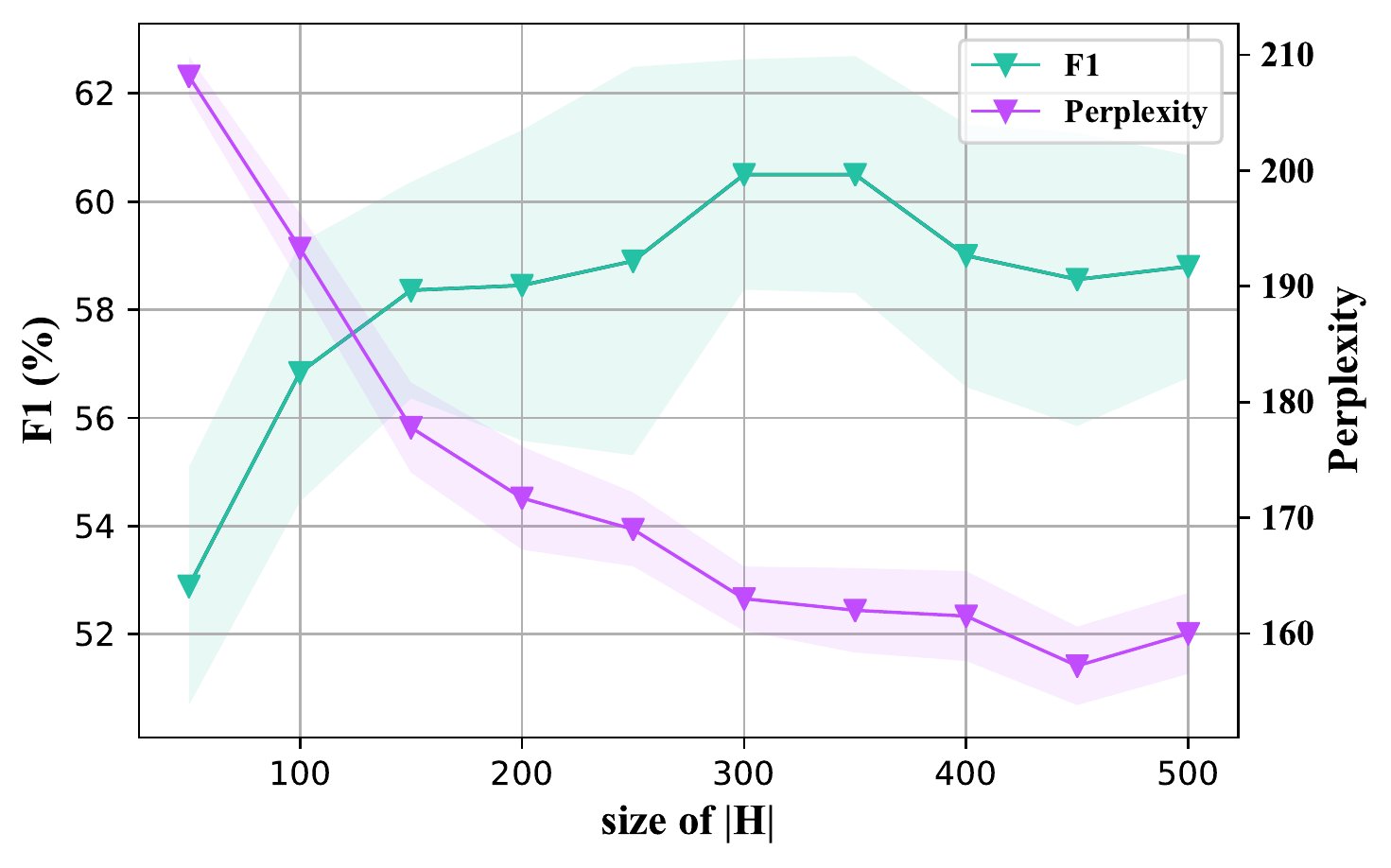}
        \caption{}
		\label{fig:f1_perp_rank}
	\end{subfigure}
	\begin{subfigure}[t]{0.49\linewidth}
		\includegraphics[scale=0.48]{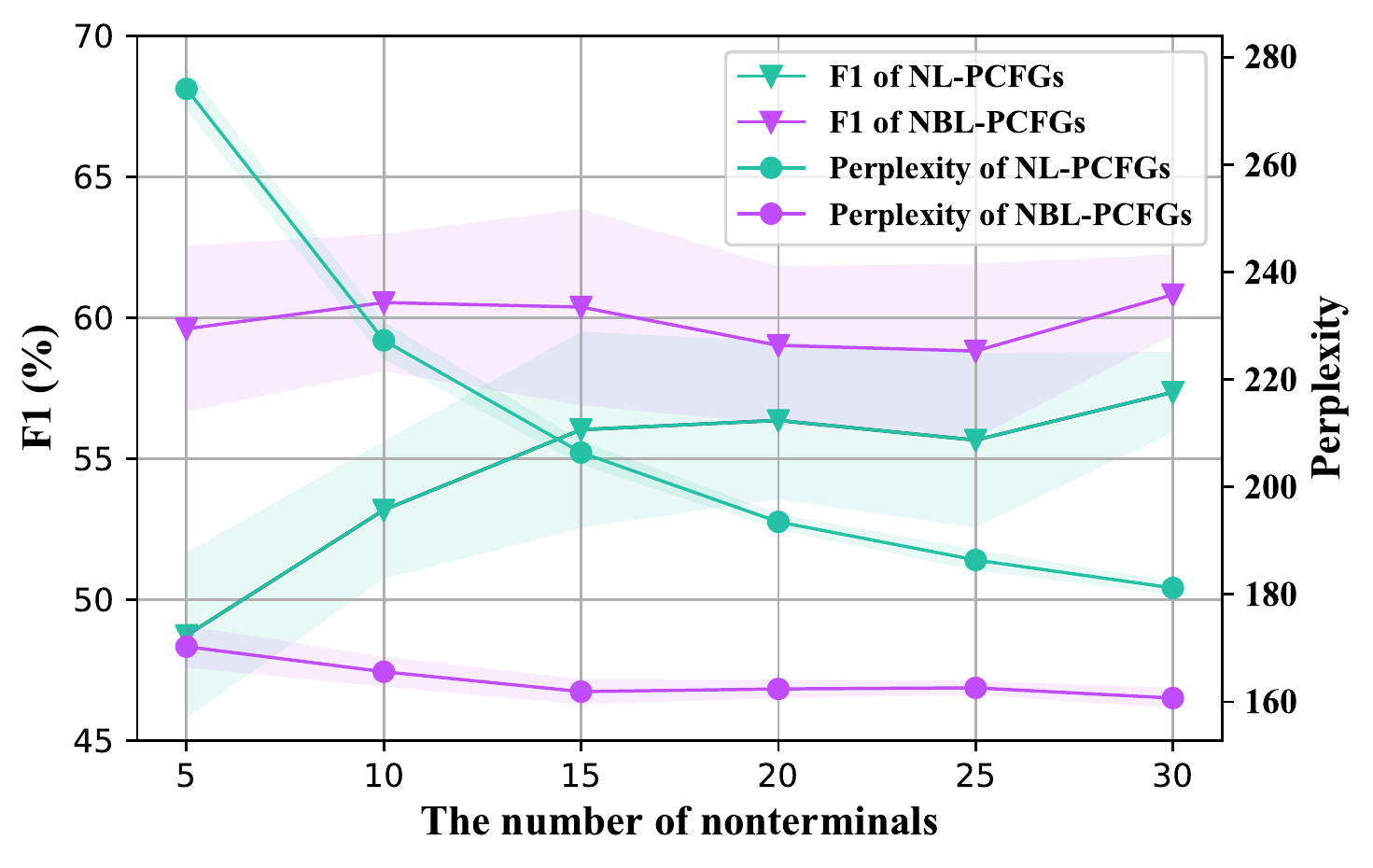}
        \caption{}
		\label{fig:f1_perp_nt}
	\end{subfigure}

	\caption{
	 The change of F1 scores,  perplexities with the change of $|H|$ and nonterminal number.
}
	\label{fig:ablation}
	\vskip -.15in
\end{figure*}

Figure \ref{fig:f1_perp_nt} illustrates perplexities and F1 scores with the increase of the nonterminal number and fixed $d_{H}=300$ (plots of UDAS and UUAS can be found in Appendix). We observe that increasing the nonterminal number has only a minor influence on NBL-PCFGs. We speculate that it is because the number of word-annotated nonterminals ($m|\Sigma|$) is already sufficiently large even if $m$ is small. On the other hand, the nonterminal number has a big influence on NL-PCFGs. This is most likely because NL-PCFGs make the independence assumption that the generation of $w_q$ is solely determined by the non-head-child $C$ and thus require more nonterminals so that $C$ has the capacity of conveying information from $A,B,D$ and $w_p$.
Using more nonterminals ($>30$) seems to be helpful for NL-PCFGs, but would be computationally too expensive due to the quadratically increased complexity in the number of nonterminals.

\subsection{Influence of different variable bindings}
\label{section:abl_bind}
Table~\ref{tab:ablating_para} presents the results of our models with the following bindings:

\begin{itemize}[leftmargin=*]\itemsep0em\setlist{nolistsep}
	\item $D$-alone: $D$ is generated alone.
	\item $D$-$w_q$: $D$ is generated with $w_q$.
	\item $D$-$B$: $D$ is generated with head-child $B$.
	\item $D$-$C$: $D$ is generated with non-head-child $C$.
\end{itemize}

Clearly, binding $D$ and $C$ (the default setting for NBL-PCFG) results in the lowest perplexity and the highest F1 score. Binding $D$ and $w_q$ has a surprisingly good performance in unsupervised dependency parsing.

We find that how to bind the head direction has a huge impact on the unsupervised parsing performance and we give the following intuition. Usually given a headword and its type, the children generated in each direction would be different. So, D is intuitively more related to $w_q$ and C than to B. On the other hand, B is dependent more on the headword instead.  In Table 3 we can see that  ($D$-$B$)  has a lower UDAS score than ($D$-$C$) and ($D$-$w_q$), which is consistent with this intuition.  Notably, in  \citet{zhu-etal-2020-return}, their Factorization III has a significantly lower UDAS than the default model (35.5 vs. 25.9), and the only difference is whether the generation of C is dependent on the head direction. This is also consistent with our intuition.


\begin{table}[t!]\small\centering
	{\setlength{\tabcolsep}{.5em}
				\begin{tabular}{lllll}
					\toprule 
				  &  F1 &  UDAS & UUAS & Perplexity \\	
				  \midrule 
				$D$-$C$ & \textbf{60.4} & 39.1 & 56.1 & \textbf{161.9} \\
				 $D$-alone & 57.2 & 32.8 & 54.1 & 164.8  \\
				  $D$-$w_q$& 47.7 & \textbf{45.7} & \textbf{58.6} & 176.8 \\
				  $D$-$B$ & 47.8 & 36.9 & 54.0 & 169.6 \\
					\bottomrule
	\end{tabular}}
	\caption{\label{tab:ablating_para} Binding the head direction $D$ with different variables.
		}
		\vskip -.1in
\end{table}

\subsection{Qualitative analysis}

We analyze the parsing performance of different PCFG extensions by breaking down their recall numbers by constituent labels (see Table \ref{tab:label_recall}). NPs and VPs cover most of the gold constituents in WSJ test set. TN-PCFGs have the best performance in predicting NPs and NBL-PCFGs have better performance in predicting other labels on average. 

We further analyze the quality of our induced trees. Our model prefers to predict left-headed constituents (i.e., constituents headed by the leftmost word). VPs are usually left-headed in English, so our model has a much higher recall on VPs and correctly predicts their headwords. SBARs often start with \emph{which} and \emph{that} and PPs often start with prepositions such as \emph{of} and \emph{for}. Our model often relies on these words to predict the correct constituents and hence erroneously predicts these words as the headwords, which hurts the dependency accuracy. For NPs, we find our model often makes mistakes in predicting adjective-noun phrases. For example, the correct parse of \emph{a rough market} is \emph{(a (rough market))}, but our model predicts \emph{((a rough) market)} instead.


\section{Discussion on dependency annotation schemes}
What should be regarded as the headwords is still debatable in linguistics, especially for those around function words \cite{Zwicky1993HeadsIG}. For example, in phrase \emph{the company}, some linguists argue that \emph{the} should be the headword \cite{Abney1972TheEN}. These disagreements are reflected in the dependency annotation schemes. 
Researchers have found that different dependency annotation schemes result in very different evaluation scores of unsupervised dependency parsing \cite{DBLP:journals/corr/Noji16, shen2020structformer}. 

In our experiments, we use the Stanford Dependencies annotation scheme in order to compare with NL-PCFGs.
Stanford Dependencies prefers to select content words as headwords.
However, as we discussed in previous sections, our model prefers to select function words (e.g., \emph{of}, \emph{which}, \emph{for}) as headwords for SBARs or PPs.
 This explains why our model can outperform all the baselines on constituency parsing but not on dependency parsing (as judged by Stanford Dependencies) at the same time. Table \ref{tab:ablating_para} shows that there is a trade-off between the F1 score and UDAS, which suggests that adapting our model to Stanford Dependencies would hurt its ability to identify constituents.



\begin{table}[t!] 	 
	{\setlength{\tabcolsep}{.5em}
		\makebox[\linewidth]{\resizebox{\linewidth}{!}{%
				\begin{tabular}{llllll}
					\toprule 
				  &  N-PCFG$^{\dagger}$ & C-PCFG$^{\dagger}$ & TN-PCFG$^{\dagger}$ & NL-PCFG &  NBL-PCFG \\	
				  \midrule				  
				 NP & 72.3\% &  73.6\% &  \textbf{75.4}\% & 74.0\% & 66.2\% \\
				 VP & 28.1\% &  45.0\% & 48.4\% & 44.3\% & \textbf{61.1}\%  \\
				 PP & 73.0\% &  71.4\% & 67.0\% & 68.4\% & \textbf{77.7}\%  \\
				 SBAR & 53.6\% & 54.8\% & 50.3\% & 49.4\% & \textbf{63.8}\% \\
				 ADJP & 40.8\% & 44.3\% & 53.6\% & 55.5\% & \textbf{59.7}\% \\
				 ADVP & 43.8\% & \textbf{61.6}\% & 59.5\% & 57.1\% & 59.1\% \\
				 \midrule
				Perplexity & 254.3 & 196.3 & 207.3 & 181.2 & \textbf{161.9} \\
				
					\bottomrule
	\end{tabular}}}}
	\caption{\label{tab:label_recall}
	Recall on six frequent constituent labels and perplexities of the WSJ test data.  $^{\dagger}$ means that the results are reported by \citet{yang-etal-2021-pcfgs}}
		\vskip -.1in
\end{table}

\section{Speed comparison}
In practice, the forward and backward pass of the inside algorithm consumes the majority of the running time in training a N(B)L-PCFG.
The existing implementation by \citet{zhu-etal-2020-return}\footnote{https://github.com/neulab/neural-lpcfg} does not employ efficient parallization and has a cubic time complexity in the number of nonterminals.
We provide an efficient reimplementation (we follow \citet{DBLP:conf/ijcai/ZhangZL20} to batchify) of the inside algorithm based on Equation \ref{eq:inside_zhu}. We refer to an implementation which caches Term C1-1 as \emph{re-impl-1} and refer to an implementation which caches Term C1-2 as \emph{re-impl-2}. 

 \begin{figure}[ht!]
	\begin{subfigure}[t]{0.49\linewidth}
		\includegraphics[scale=0.26]{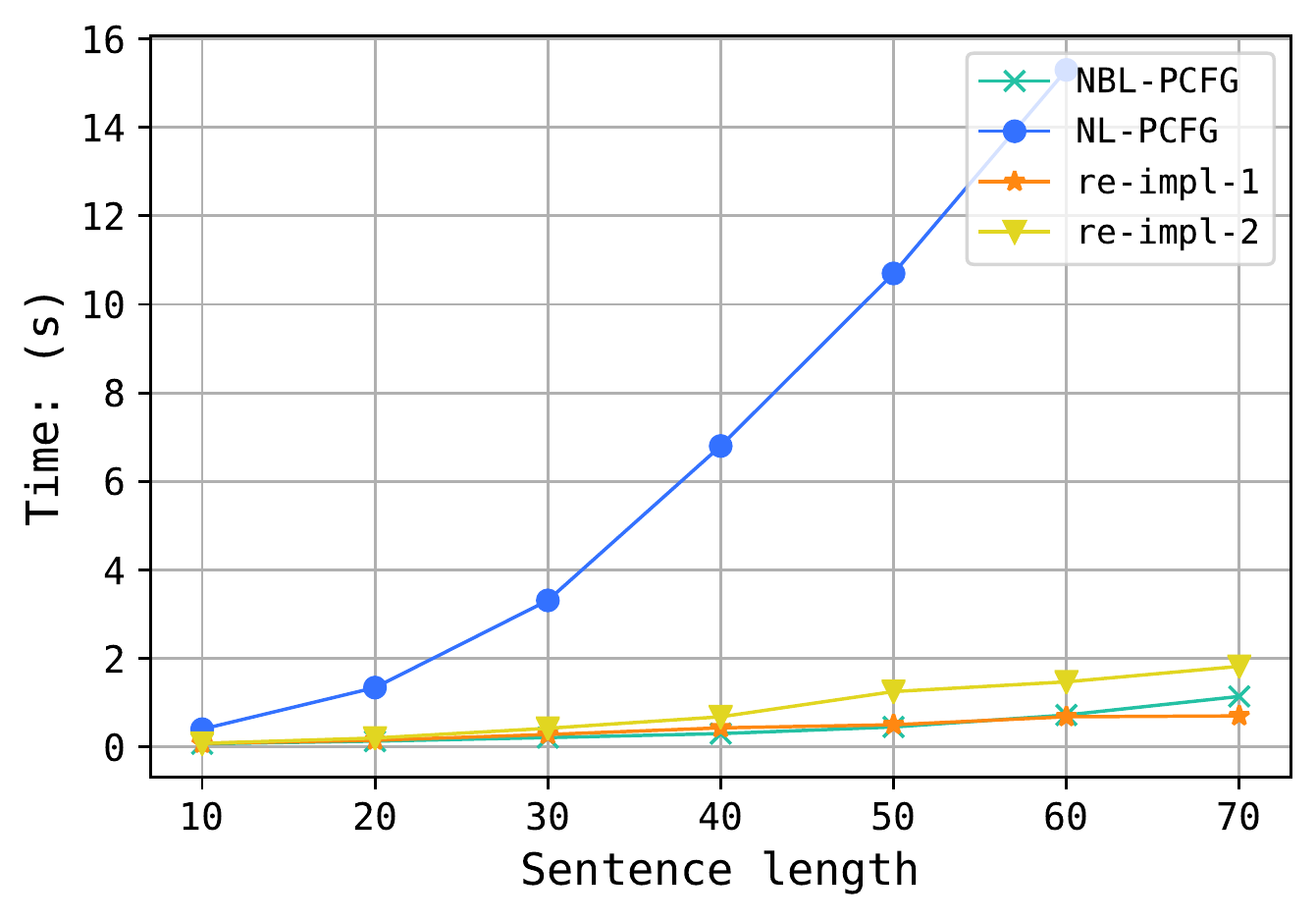}
        \caption{}
		\label{fig:time_length}
	\end{subfigure}
	\begin{subfigure}[t]{0.49\linewidth}
		\includegraphics[scale=0.26]{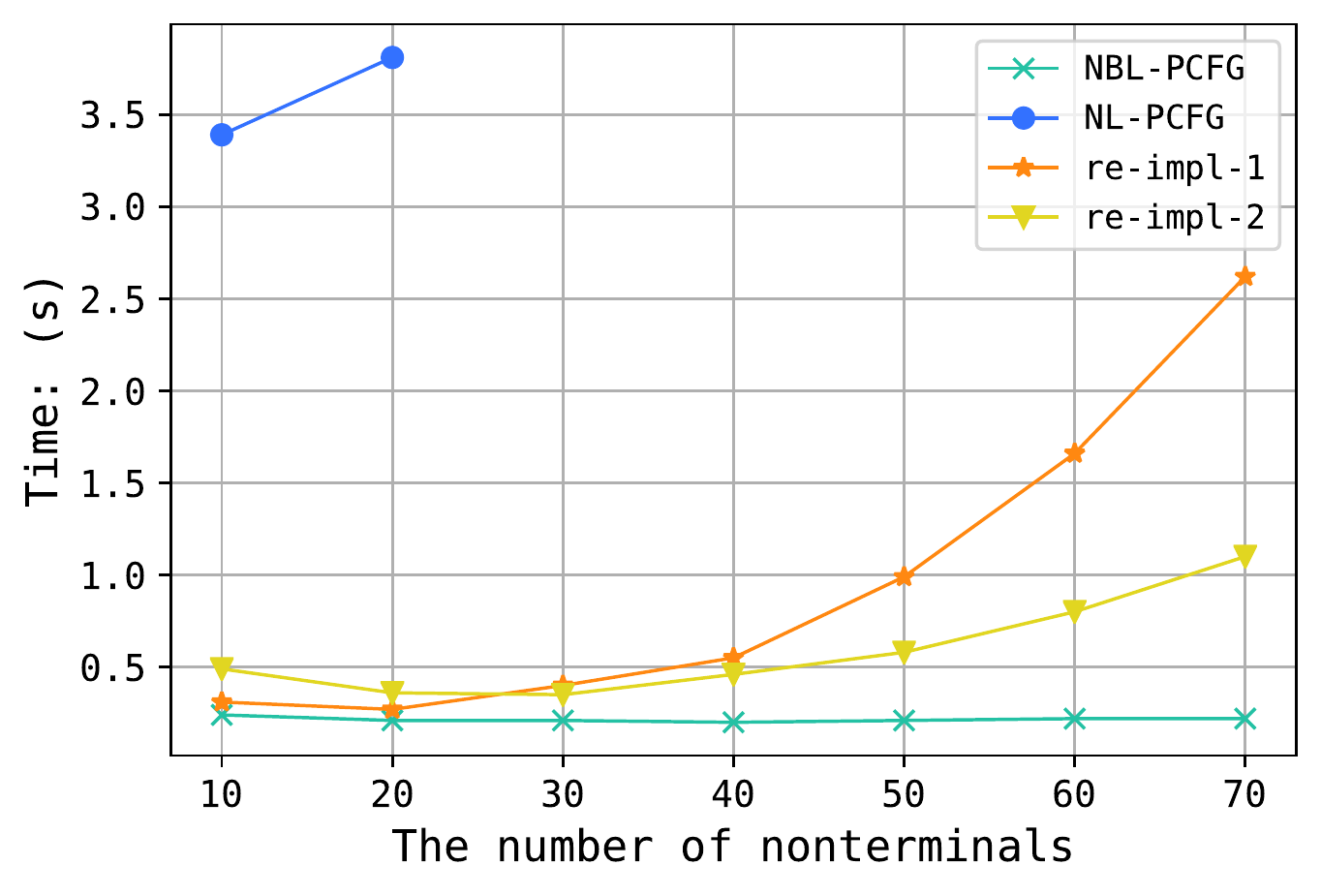}
        \caption{}
		\label{fig:time_nt}
	\end{subfigure}

	\caption{
	 Total time in performing the inside algorithm and automatic differentiation with different sentence lengths and nonterminal numbers. 
}
	\label{fig:time}
	\vskip -.15in
\end{figure}
 
We measure the time based on a single forward and backward pass of the inside algorithm with batch size 1 on a single Titan V GPU.  Figure \ref{fig:time_length} illustrates the time with the increase of the sentence length and a fixed nonterminal number of 10. The original implementation of NL-PCFG by \citet{zhu-etal-2020-return} takes much more time when sentences are long. For example, when sentence length is 40, it needs 6.80s, while our fast implementation takes 0.43s and our NBL-PCFG takes only 0.30s. Figure \ref{fig:time_nt} illustrates the time with the increase of the nonterminal number $m$ and a fixed sentence length of 30. The original implementation runs out of 12GB memory when $m=30$. \emph{re-impl-2} is faster than \emph{re-impl-1} when increasing $m$ as it has a better time complexity in $m$ (quadratic for \emph{re-impl-2}, cubic for \emph{re-impl-1}). Our NBL-PCFGs have a linear complexity in $m$, and as we can see in the figure, our NBL-PCFGs are much faster when $m$ is large.

\section{Related Work}
Unsupervised parsing has a long history but has regained great attention in recent years. In unsupervised dependency parsing, most methods are based on Dependency Model with Valence (DMV) \cite{klein-manning-2004-corpus}. Neurally parameterized DMVs have obtained state-of-the-art performance \cite{jiang-etal-2016-unsupervised, han-etal-2017-dependency, han-etal-2019-enhancing, yang-etal-2020-second}. However, they rely on gold POS tags and sophisticated initializations (e.g. K\&M initialization or initialization with the parsing result of another unsupervised model). \citet{noji-etal-2016-using} propose a left-corner parsing-based DMV model to limit the stack depth of center-embedding, which is insensitive to initialization but needs gold POS tags. \citet{he-etal-2018-unsupervised} propose a latent-variable based DMV model, which does not need gold POS tags but requires good initialization and high-quality induced POS tags. See \citet{han-etal-2020-survey} for a survey of unsupervised dependency parsing.
Compared to these methods, our method does not require gold/induced POS tags or sophisticated initializations, though its performance lags behind some of these previous methods.  

Recent unsupervised constituency parsers can be roughly categorized into the following groups: (1) PCFG-based methods. Depth-bounded PCFGs \cite{jin-etal-2018-depth,jin-etal-2018-unsupervised} limit the stack depth of center-embedding. Neurally parameterized PCFGs \cite{jin-etal-2019-unsupervised, kim-etal-2019-compound, zhu-etal-2020-return, yang-etal-2021-pcfgs} use neural networks to produce grammar rule probabilities. (2) Deep Inside-Outside Recursive Auto-encoder (DIORA) based methods \cite{drozdov-etal-2019-unsupervised-latent, drozdov-etal-2019-unsupervised, drozdov-etal-2020-unsupervised, hong-etal-2020-deep, sahay2021rule}. They use neural networks to mimic the inside-outside algorithm and they are trained with masked language model objectives. (3) Syntactic distance-based methods \cite{DBLP:conf/iclr/ShenLHC18, DBLP:journals/corr/abs-1910-13466,  shen2020structformer}. They encode hidden syntactic trees into syntactic distances and inject them into language models. (4) Probing based methods \cite{DBLP:conf/iclr/KimCEL20, li-etal-2020-heads}.  They extract 
phrase-structure trees based on the attention distributions of large pre-trained language models.
In addition to these methods, \citet{cao-etal-2020-unsupervised-parsing} use constituency tests and \citet{shi-etal-2021-learning} make use of naturally-occurring bracketings such as hyperlinks on webpages to train parsers. Multimodal information such as images \cite{shi-etal-2019-visually, zhao-titov-2020-visually, jin-schuler-2020-grounded} and videos \cite{zhang-etal-2021-video} have also been exploited for unsupervised constituency parsing. 

We are only aware of a few previous studies in unsupervised joint dependency and constituency parsing. \citet{klein-manning-2004-corpus} propose a joint DMV and CCM \cite{klein-manning-2002-generative} model. \citet{shen2020structformer} propose a transformer-based method, in which they define syntactic distances to guild attentions of transformers. \citet{zhu-etal-2020-return} propose neural L-PCFGs for unsupervised joint parsing.

\section{Conclusion}
We have presented a new formalism of lexicalized PCFGs.
Our formalism relies on the canonical polyadic decomposition to factorize the probability tensor of binary rules.
The factorization reduces the space and time complexity of lexicalized PCFGs while keeping the independence assumptions encoded in the original binary rules intact.
We further parameterize our model by using neural networks and present an efficient implementation of our model.
On the English WSJ test data,
our model achieves the lowest perplexity,
outperforms all the existing extensions of PCFGs in constituency grammar induction,
and is comparable to strong baselines in dependency grammar induction.

\section*{Acknowledgments}
We thank the anonymous reviewers for their constructive comments. This work was supported by the National Natural Science Foundation of China (61976139).

\bibliographystyle{acl_natbib}
\bibliography{anthology,acl2021}

\appendix
\section{Full Parameterization}
We give the full parameterizations of the following probability distributions.

\begin{align*}
p(  A|S) &= \frac{\exp (\mathbf{u}_{S}^{\top} h_1(\mb{w}_{A}))}{  \sum_{A^{\prime} \in \mathcal{N}} \exp (\mathbf{u}_{S}^{\top} h_1(\mb{w}_{A^{\prime}})) } \,,\\
p( w|A) &= \frac{\exp (\mathbf{u}_{A}^{\top} h_{2}( \mb{w}_{w}))}{  \sum_{w^{\prime} \in \Sigma} \exp (\mathbf{u}_{A}^{\top} h_{2}(\mb{w}_{w'}))}\,, \\
p(w |H) &= \frac{\exp (\mathbf{u}_{H}^{\top} h_3 ( \mb{w}_{w}) )}{  \sum_{w^{\prime} \in \Sigma} \exp (\mathbf{u}_{H}^{\top} h_3 ( \mb{w}_{w^{\prime}})) } \,, \\
p(H|A, w ) &=  \frac{ \exp (\mathbf{u}_{H}^{\top} f([\mathbf{w}_{A}; \mathbf{w}_{w}]) ) }{\sum_{H^{\prime} \in \mathcal{H}} \exp (\mathbf{u}_{h^{\prime}}^{\top} f([\mathbf{w}_{A}; \mathbf{w}_{w}])) } \,,
\end{align*}

$\begin{aligned} h_{i}(\mathbf{x}) &=g_{i, 1}\left(g_{i, 2}\left(\mathbf{W}_{i} \mathbf{x}\right)\right) \\ g_{i, j}(\mathbf{y}) &=\operatorname{ReLU}\left(\mathbf{V}_{i, j} \operatorname{ReLU}\left(\mathbf{U}_{i, j} \mathbf{y}\right)\right)+\mathbf{y}\\  f([\mathbf{x}, \mathbf{y}]) &=  h_{4}( \operatorname{ReLU}(\mathbf{W}[x;y]) + y)\\
\end{aligned}
$

\begin{figure}[tb!]
	\centering
	\includegraphics[width=.9\linewidth]{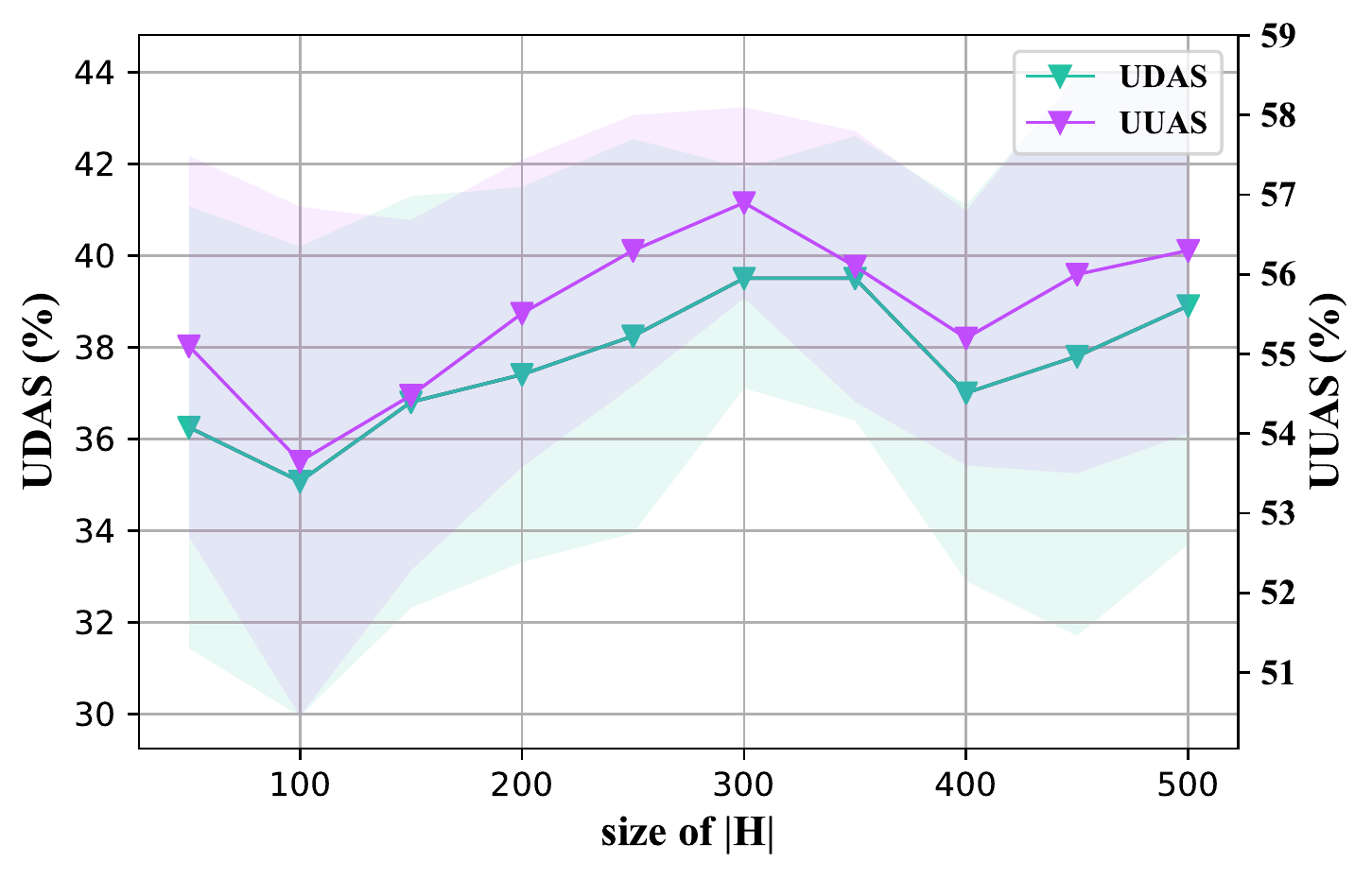}
	\caption{Influence of $d_H$ on  UUAS and UDAS. }
	\label{fig:das_h}
\end{figure}
\begin{figure}[tb!]
	\centering
	\includegraphics[width=.9\linewidth]{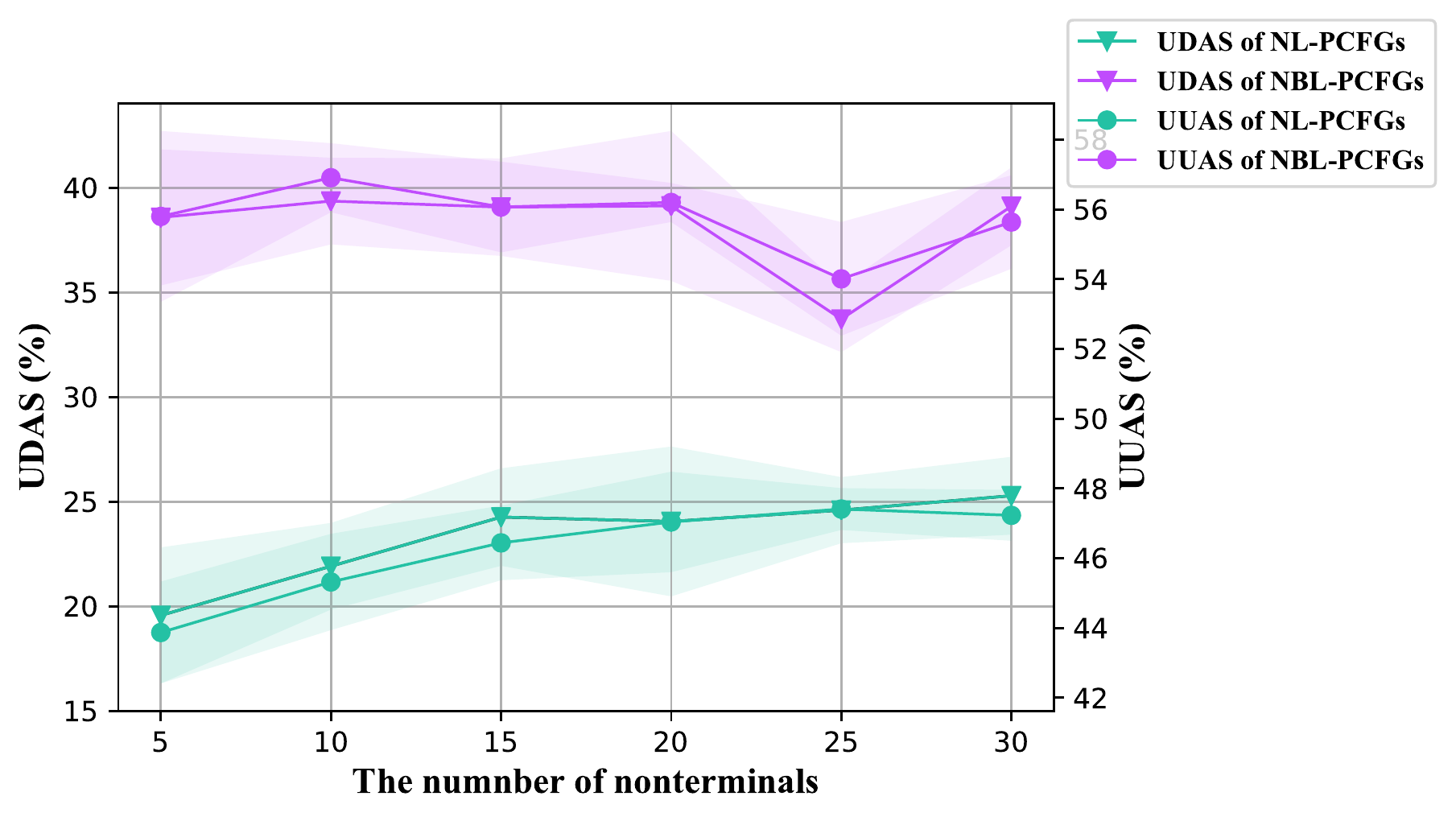}
	\caption{Influence of the number of nonterminals on UUAS and UDAS.}
	\label{fig:das_nt}
\end{figure}

\section{Influence of the domain size of $H$ and the number of nonterminals }
Figure \ref{fig:das_h} illustrates the change of UUAS and UDAS with the increase of $d_H$. We find similar tendencies compared to the change of F1 scores and perplexities with the increase of $d_H$. $d_H = 300$ performs best. Figure \ref{fig:das_nt} illustrates the change of UUAS and UDAS when increasing the number of nonterminals. We can see that NL-PCFGs benefit from using more nonterminals while NBL-PCFGs have a better performance when the number of nonterminals is relatively small.




\end{document}